# Key Phrase Extraction & Applause Prediction

*Auto ML tool for performing ML techniques on json files of medium online article dataset

1st Krishna Yadav
*Computer Science Department (IIITD)*
*Indraprastha Institute of Information Technology (AICTE)*
Delhi, India
krishna19039@iiitd.ac.in

2nd Lakshya Choudhary
*Computer Science Department (IIITD)*
*Indraprastha Institute of Information Technology (AICTE)*
Delhi, India
lakshya19067@iiitd.ac.in

*Abstract* – With increase in content availability over the internet it is very difficult to get noticed. It has become an upmost priority of the blog writers to get some feedback over their creations to be confident about the impact of their article. We are training a machine learning model to learn popular article styles, in the form of vector space representations using various word embeddings, and their popularity based on claps and tags.

## 1. Problem Statement and Motivation

Today's world is full of online content and lot more is producing each day. With increase in content availability over the internet it is very difficult to get noticed.

Medium is one of the top platforms for spreading information and knowledge about almost any domain in the society. It is abundantly used to publish content, articles on computer science domain such as ML, AI, Data Engineering, etc.

## 2. Introduction

Writing articles that can get high popularity is a difficult task. The problem comes for the new authors who don't have much idea about how to get popularity. We have developed intuition into making an interesting and interesting Medium article. Instead of purely relying on a guess theory, we wanted to test our theory using data science and machine learning. As a Data Scientist, we have always been curious about creativity and automation Also, believe that our article and hard work on this model combines both faces of science to predict the best Medium article. We introduce our work and further work Platforms like Medium have their own metrics to evaluate the content by allowing readers to clap to show their appreciation. Also, these platforms offer provision to add few Key-Phrases that best provides the subject of the content.

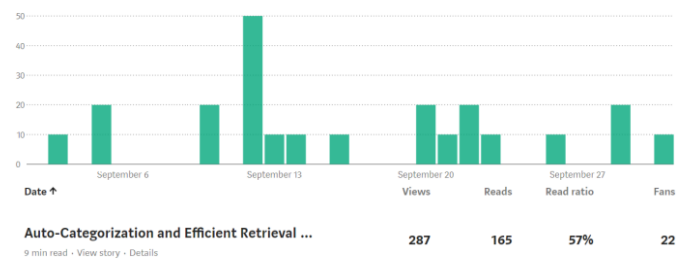

*Figure 1 Medium Article Statistics*

Now, it becomes a necessity of the writer to get some assistance for best possible key-phrases to get the articles ranked up and tentative popularity of the new article. Our aim in this project will be to train a model on the Medium.

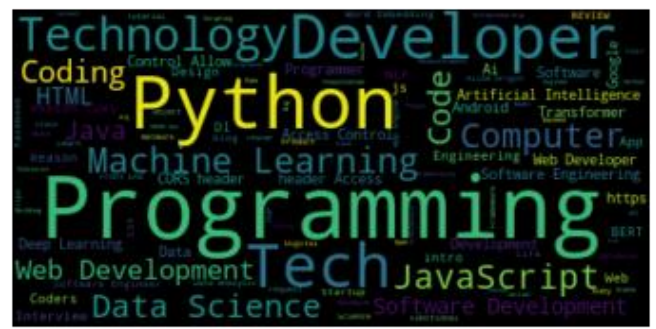

*Figure 2 Word Cloud: Extracting tags from articles*

## 3. Literature Survey

Key phrase extraction is a big field and a lot of people have worked on it already. We have many pre-trained models like Bert and pke rake etc. that helps to extract the key phrases. Although we have many advancements in the text classification, this whole problem of article performance problem is explored by only a smaller group of researchers. pke is an open source python-based key phrase extraction toolkit. It provides an end-to-end key phrase extraction pipeline in which each component can be easily modified or extended to develop new models. pke also allows for easy benchmarking of state-of-the-art key phrase extraction models, and ships with supervised models trained on the SemEval-2010 dataset. [1], [2]. Sahrawat et al. [8] have used neural network models such as bert and bi-lstm to extract key phrases from scholarly articles as a sequence labelling task solved using a BiLSTM-CRF, where the words in the input text are represented using deep contextualized embeddings.

## 4. Dataset Used

The dataset that we used for this project is being scrapped using selenium in the form of Json strings from the official Medium website. So far, we have over 2574 articles of diverse popularity. We are collecting more data but since the annotation verification takes time so will conclude this in final presentation. For pre-processing we removed all the Non_ASCII characters and the white-spaces or special characters. then we converted the digits into their English equivalent. followed by spell correction and lemmatization.

```
{'applause': 0,
 'author': 'DataLabeler L',
 'id': 'fff11c791c5a',
 'publish_date': '2020-01-08',
 'reading_time': '3',
 'tags': ['Machine Learning',
  'Data Labeling',
  'Datalabeler',
  'Data Labeling Service',
  'Dl'],
 'title': 'NLP & its Uses | Da
 'txt': 'Natural Language Proc
}
```

Our data had a lot of new articles having zero claps so we removed those to avoid misdirection. and finally took doc2vec embeddings to get a numerical document vector.

## 5. Data Scrapping

Web Scrapping also called "Crawling" or "Spidering" is the technique to gather data automatically from an online source usually from a website. While Web Scrapping is an easy way to get a large volume of data in a relatively short time frame, it adds stress to the server where the source is hosted. We scrapped the data from medium using selenium. Data mining or gathering data is a very primitive step in the data science life cycle. As per business requirements, one may have to gather data from sources like SAP servers, logs, Databases, APIs, online repositories, or web. Tools for web scraping like Selenium can scrape a large volume of data such as text and images in a relatively short time.

Selenium is an open-source web-based automation tool. Selenium primarily used for testing in the industry but It can also be used for web scraping. We'll use the Chrome browser but you can try on any browser, It's almost the same. [5] We scrapped the by following steps:
- Search URLs of articles with the given set of tags to capture domain related articles only.
- Scrapping the useful data from each webpage.
- Saving the scraped article data in json object.

## 6. Pre-processing

The steps for natural language pre-processing are taken up to convert each sentence into equivalent processed word representation [1], [2], [7], [8].

The techniques used to process each word are:

i. Punctuation Removal: *'.",?~!<@#$%&*)*
ii. Whitespaces Removal: *\n \t \a*
iii. Non-Ascii Characters: *non-keyboard, emoji*
iv. Tokenization: *'hello', 'world'*
v. Number conversion: *1* to *'one'*
vi. Stop-word removal: *the, are, is*
vii. Spelling correction: *thro, teling, forgt*
viii. Word segmentation: *haveto, whynot*
ix. Lemmatization: *mice* to *mouse*
x. Expanding abbreviations: *I've, can't*
xi. Scaling: *[1, 90,99]* to *[0.1, 0.9, 0.99]*

Text data is not suitable to feed into ML models, so we used Glove and doc2vec combined with sequences obtained by the text_to_sequence method from *Keras*' tokenizer.

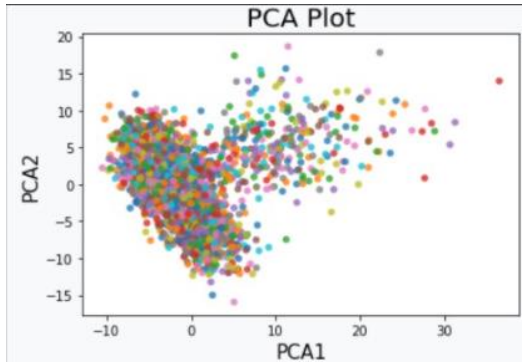

*Figure 3 Data Distribution in 2D Space*

## 7. Feature Extraction

After generalizing the length of the questions, we are left with long vectors with a lot of relevant information (be it more or less). Since the length is directly proportional to the vocab size, as mention in *Zip's law* [10], [11], it is crucial to extract interesting features only. We tried PCA and LDA feature extraction[12]. Word embeddings also helped and finally got 23 top features to apply ML algorithms. More or fewer features than 23 lead us to the reduced performance of our model. That's why we are only capturing the most relevant information with this set of features only. Dimensionality reduction also helps to visualize the top data components for better understanding, as shown in Figure 3.

## 8. Proposed Model

We tried several approaches and combinations to reach to the final set of conclusions. However, the pre-processing steps for all remain the same. Pre-processing differs for regression task and keyword prediction task.

i. <u>Baseline Models</u>

We have used Liner Regression for regression over the claps and we also proposed the use of KNN to find the closest similar data points to our test document in vector space. Regression gives MSE i.e. we calculated the deviation of the predicted results from actual ground truth results. This is to get the factor of claps at each document.

ii. <u>Deep Learning Models</u>

Deep Learning has been proven to solve many hard problems. We have planned to use deeply learned models like RoBERTa to get optimised results for our model final prediction for the article being popular or not. We are planning to use BIO-Tagging to represent each document for key phrase extraction. It is a common format the similar kind of NLP tasks such as NER. And later using deep neural network to predict those BIO tags as labels.

| | |
|---|---|
| at | O |
| the | O |
| United | B-org |
| Nations | I-org |
| summit | O |
| in | O |
| New | B-geo |
| York | I-geo |
| , | O |
| Prime | B-per |
| Minister | O |
| Fouad | B-per |
| Siniora | I-per |
| said | O |

## 9. Performance Improvement

*What Improved our Performance, and why?*

i. Glove: performs better than doc2vec as it captures sublinear relationships.
ii. Stop-words: In our case, stop words are worthless as simply misdirect the classification as they occur very often in every document.
iii. Dimension Reduction works: our vector representation has a lot of features and feeding that to a model increases time and hides essential information.
iv. Adding features: Selected word embeddings give us 23 sufficient features. Adding or removing features reduces the performance as the relevant information gets compromised.

*What didn't work, and why?*

i. Spelling Correction: Most of the questions didn't have any spelling mistakes. It is assumed that the user is aware of the language.
ii. Scaling: Vector representation of questions is already scaled. Further scaling doesn't help anymore.
iii. All embeddings: Embeddings usually capture most of the information. Combining genism embeddings doesn't solve the purpose.

- iv. Adding features: Selected word embeddings give us 23 sufficient features. Adding or removing features reduces the performance as the relevant information gets compromised.
- v. Reducing Vector size: It improves the training time but at the cost reduced performance.

## 10. Baseline Results

For regression task we used linear regression to draw a linear decision boundary and estimate the claps of new articles. Also, we propose the use of knn in terms of document similarity to find k nearest documents to the new document in multi-dimension vector space and take average of those document claps. Tuning the hyperparameters showed a lot of improvement. We are planning task specific fine tuning for both of these tasks.

| Model | MSE |
|---|---|
| LR + Doc2Vec(Embedding Size 100) | 12431 |
| LR + Doc2Vec(Embedding Size 200) | 6131 |
| KNN + Doc2Vec(Embedding 200, k=19) | 3054 |

*Table 1 Applause Prediction: Baseline Models*

For Key-Phrase Extraction task, we initialized Topic-Rank key phrase extraction model and carried the preprocessing using spacy. The key phrase candidate selection happens as sequences of nouns and adjectives (i.e. `(Noun|Adj)*`). Then candidate weighting happens using a random walk algorithm. And finally, N-best key phrases i.e. highest scored candidates are extracted as (keyphrase, score) tuples.

```
developer
tech decades
tech industry
internet
new skills
demand
new career
today
personal computers
careers
```

*Table 2 Predicted Key Phrases*

```
Front End Web Developer
Full Stack JavaScript Developer
Java
Java Web Developer
Python Web Developer
versatile
Python
Nasa
Android Developer
iOS Developer
Tech
Web Development
Web Developer
Coding
Programming
```

*Table 3 Actual Key Phrases*

## 11. Future Work Conclusion

We proposed KNN to predict number of claps. Further we plan to incorporate other features with more weightage i.e. date of publish, author name etc. We will use deep learning and models like Bert to improve the performance of key phrase extraction task. We are planning to use high, medium, low to measures to count the claps and then perform classification using clap and the old features as parameters. We expect a drastic performance increase. Increasing data corpus size.

## 12. Authors' Contributions

- i. Krishna Yadav (MT19039)
    - a. Data cleaning and Pre-processing
    - b. Data Annotation
    - c. Data Analysis and visualization
    - d. Key Phrase Extraction
    - e. Baseline Results

- ii. Lakshya Choudhary (MT19067)
    - a. Data Crawler
    - b. Data Annotation
    - c. Feature selection
    - d. LR, KNN | Parameter Tuning
    - e. Baseline Results